# Deep Learning for Hydroelectric Optimization: Generating Long-Term River Discharge Scenarios with Ensemble Forecasts from Global Circulation Models

Julio A. S. Dias (alberto@psr-inc.com)

PSR Energy Consulting

*Abstract*-- **Hydroelectric power generation stands as a critical component of the global energy matrix, particularly in countries like Brazil, where it constitutes most of the energy supply. However, this type of generation is highly dependent on river discharges, which are intrinsically uncertain due to the variability of the climate, as river flows are directly linked to precipitation volumes. Therefore, the development of accurate probabilistic forecasting models for river discharges is crucial to enhance the operational planning of systems heavily reliant on this resource.**

**Traditionally, the representation of river discharges in energy resource optimization applications has relied on statistical models. However, these models have become increasingly inadequate in generating realistic scenarios, primarily due to structural shifts in climate behavior. Changes in precipitation patterns have led to alterations in discharge patterns, which are not accurately captured by these traditional statistical approaches. On the other hand, machine learning methods, which have proven effective as universal predictors for time series data, typically focus on making predictions based solely on the historical time series itself, often overlooking critical external factors. In river discharge scenario generation, this approach falls short for several reasons: firstly, it overlooks valuable information by disregarding current meteorological and climatic conditions; secondly, the inherent variability of hydrological processes calls for a probabilistic approach to this specific problem; and thirdly, the limited availability of historical data contrasts sharply with the vast datasets typically used to train large-scale deep learning models.**

**In order to take advantage of the potential of deep learning architectures while addressing the practical challenges of the specific application, we propose a framework based on a modified recurrent neural network architecture. This architecture not only generates parameterized probability distributions, conditioned on projections from global circulation models, effectively managing the stochastic nature of the problem, but also integrates modifications that enhance its generalization capability.**

**We implement this model within the Brazilian Interconnected System, integrating seasonal projections from the SEAS5-ECMWF system as conditional variables. The results demonstrate significant improvements in forecast accuracy, enhancing the generation of more realistic future discharge scenarios, which directly and positively influence operational planning strategies.**

## I. Introduction

Hydroelectric power generation plays a central role in the global energy matrix, particularly in countries like Brazil, where it constitutes a majority of the energy supply. However, this form of energy generation is highly dependent on river discharges, which are subject to inherent uncertainty due to the stochastic nature of hydrological processes. These discharges are directly influenced by precipitation patterns, which can be significantly altered by changing climate conditions. These alterations may arise from cyclical phenomena such as El Niño and La Niña, which cause periodic shifts in weather patterns, including changes in rainfall distribution and intensity [1],[2],[3]. Additionally, more structural changes driven by long-term climate change can profoundly affect precipitation patterns [4],[5]. As a result, the ability to predict river discharge has become a critical challenge for the efficient operation and planning of hydroelectric systems. Traditionally, the generation of future river discharge scenarios for energy resource optimization has relied on statistical regression models, such as [6],[7], which have become increasingly inadequate. These models struggle to handle the complexity of climate behavior, particularly as shifts in precipitation patterns lead to changes in river discharge patterns that cannot be effectively reproduced by conventional statistical methods.

In recent years, machine learning (ML) techniques have emerged as powerful tools for predictive modeling, particularly in time series forecasting. While ML methods, such as deep learning models, have achieved remarkable success across various domains, they often rely solely on historical data to make predictions, overlooking critical external drivers like climate dynamics. For river discharge forecasting, this limitation is increasingly problematic. Shifts in climatic patterns have made it challenging to anticipate hydrological behavior based solely on past observations, as historical data may not fully reflect the evolving climatic conditions that influence river discharge. Another challenge lies in the inherent variability of hydrological processes, which necessitates the use of a probabilistic approach. Such an approach is essential to account not only for the uncertainty in future river discharge but also for the influence of external conditioning variables that drive this variability. In this context, estimating probability distributions directly from the time series data offers a more accurate representation of uncertainty compared to traditional methods that build predictive models and model uncertainty through the stochastic representation of residuals. While regression-based models may struggle to capture the time-varying nature of hydrological systems or require complex adjustments to handle non-stationary data, direct distribution estimation provides a more flexible and robust framework.

To address these challenges, we propose a novel deep learning framework based on a modified Gated Recurrent Unit (GRU) architecture for river discharge scenario generation. Unlike traditional machine learning models, which predict time series based solely on historical data, our approach integrates seasonal weather forecasts derived from Global Circulation Models (GCMs), providing a more comprehensive and accurate representation of the factors influencing river discharge. By incorporating weather ensembles into the model, we can generate probabilistic forecasts that account for the inherent uncertainty of future discharge, rather than offering deterministic predictions that may be overly confident or inaccurate.

The new architecture was designed to address several key challenges that have traditionally limited the effectiveness of ML models in this context. First, the model is capable of generating parameterized probability distributions, which allows for a more robust representation of uncertainty compared to point estimates. Second, it integrates information from GCMs, ensuring that a diverse set of weather scenarios is considered in the forecasting process. This is particularly important for long-term predictions, as climate variability can lead to significant changes in discharge patterns over time. Finally, the architecture's design incorporates mechanisms to prevent overtraining, a common issue in deep learning applications where models become overly fitted to the available data and lose their ability to generalize to new, unseen scenarios. By addressing these challenges, our framework offers a more accurate and reliable approach for generating long-term river discharge scenarios that can inform decision-making in hydroelectric optimization and planning.

The proposed model is applied to the Brazilian Interconnected System. The results show substantial improvements in forecast accuracy, generating more realistic future discharge scenarios and providing valuable insights for operational planning. By incorporating external weather forecast data into the model, we can account for a wide range of potential climate conditions, allowing hydroelectric systems to better anticipate and respond to future changes in river discharge.

In the following sections, we will detail the methodology behind the proposed deep learning framework, including the integration of weather ensembles, the modification of the recurrent network architecture, and the application of the model to the Brazilian Interconnected System. We will also present the results of our experiments, demonstrating the framework's effectiveness in generating more realistic river discharge scenarios and enhancing the accuracy of medium- and long-term forecasting.

It is important to emphasize that the objective of this work is not to evaluate the performance of different GCMs in representing the climate of various regions in Brazil. While this is a critical aspect for enhancing climate-related applications, it lies beyond the scope of this study, which focuses on the development and validation of the proposed model. In this work, a single GCM was selected based on its performance reported in recent literature. Future studies will build upon the methodologies and resources presented here to thoroughly assess and identify the GCMs that best represent Brazil's regional climates, enabling a more robust and accurate approach to scenario generation.

## II. PROPOSED ARCHITETURE

The proposed architecture is designed to generate scenarios for each hydrographic basin based on the external driver provided by the ensemble of climate projections. For each basin, the input consists of precipitation and temperature projections provided in grid form, corresponding to the spatial distribution over the basin area. These grid-based projections provided by GCMs, serve as the external driver for the model, which then generates discharge scenarios for each hydroelectric plant located within the basin.

Fig.1 illustrates the hydrographic basins of Brazil, with each basin represented by areas shaded in different tones of blue. The intensity of the blue shading indicates the relative average generation of the hydroelectric plants within the basin—darker shades correspond to higher generation levels, while lighter shades represent lower generation. The red dots on the map mark the locations of the hydroelectric plants within these basins, highlighting the areas of interest for the scenario generation process.

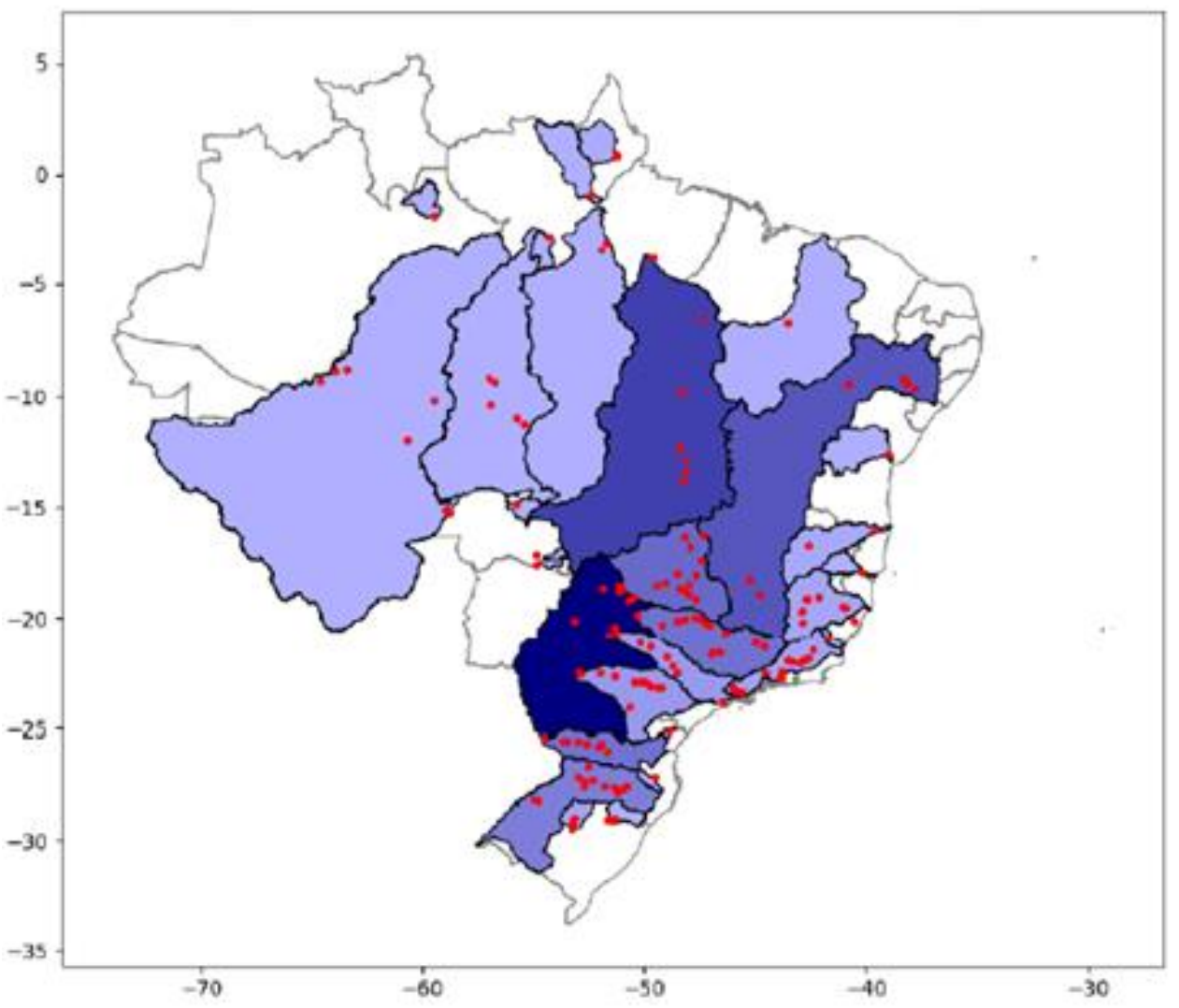

Fig.1. Brazil Hydro Basins and Hydroelectric Plants

As will be further detailed in the following subsections, the architecture can be seen as a mapping from sequences of grid-based precipitation and temperature projections to joint probability distributions of river discharges. One notable advantage of this proposed mapping is its ability to generate an unlimited number of river discharge scenarios from a finite set of ensemble projections, enabling the exploration of a broader range of potential hydrological outcomes while maintaining consistency with the base climate projections.

A schematic view is presented in Fig.2, where the blue, red, and green lines represent three trajectories (ensembles) projected by the GCM model, which feed into the proposed model, generating a set of hydrology scenarios for each ensemble.

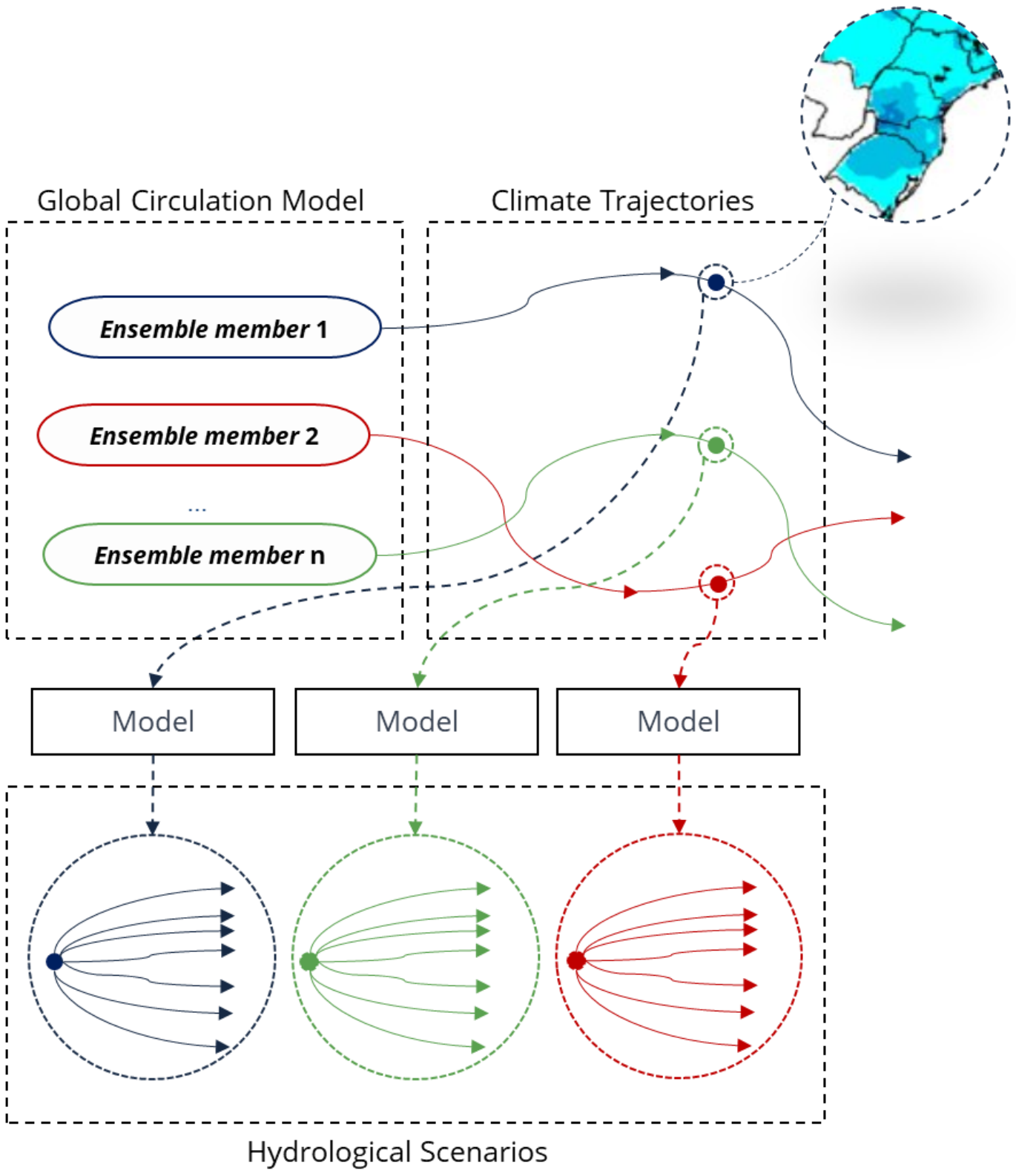

Fig.2. Hydrological Scenarios Generation from GCM Ensembles

## A. Gated Recurrent Layer Architecture

Gated Recurrent Units[8] (GRUs) are a type of recurrent neural network (RNN) architecture designed to address challenges in modeling sequential data. Unlike traditional RNNs, GRUs use gating mechanisms to control the flow of information through the network, making them more effective at capturing temporal patterns. Specifically, GRUs consist of two key gates: the reset gate and the update gate. The reset gate determines how much of the previous information should be forgotten, while the update gate controls how much of the new information should be added to the current state.

Although the concepts of forgetting and updating information of the hidden state are often associated with the context in natural language processing applications, these ideas are equally applicable to the representation of time-dependent dynamics, such as those encountered in hydrology. In hydrological modeling, for example, the reset and update gates can be interpreted as mechanisms for managing the filling and draining of reservoirs, both on the surface and underground. Just as the model learns which past information is relevant to predict the next word in a sentence, the GRU can determine which past hydrological states (e.g., soil moisture levels, groundwater storage) should be retained or updated to best predict future river discharge.

By effectively retaining or discarding information, GRUs can model complex patterns in sequential data, such as seasonality. Their relatively simple architecture, compared to alternatives like Long Short-Term Memory (LSTM) networks, not only makes GRUs computationally efficient—reducing training time without compromising performance—but also aligns well with scenarios where data availability is limited. This simplicity enables GRUs to perform robustly in applications with fewer data points, addressing challenges that are typically more demanding for recurrent networks with more complex structures.

The typical architecture of a GRU layer is presented in Fig.3, while the applied formulation is presented in (1).

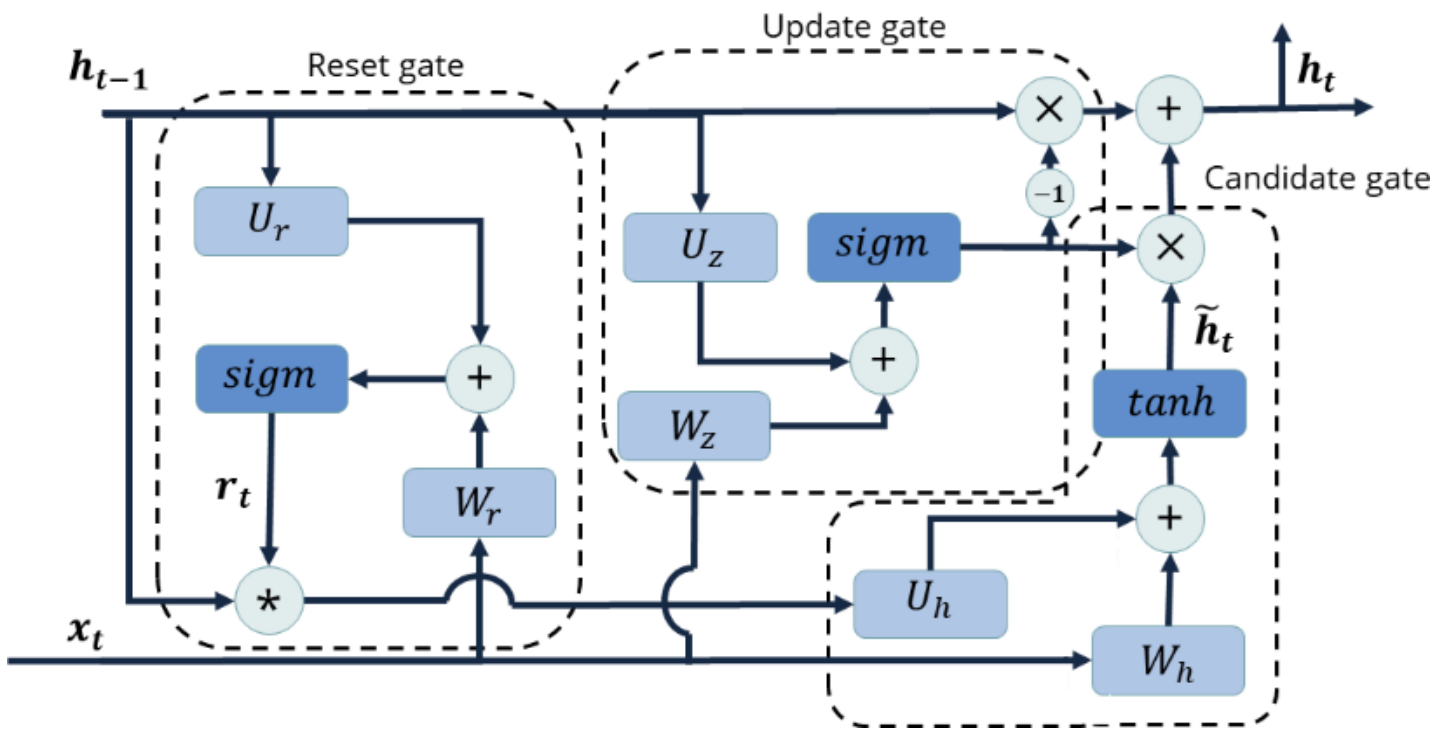

Fig.3. Gated Recurrent Unit layer

$$
\begin{aligned}
z_t &= sigm(W_z \cdot x_t + U_z \cdot h_{t-1}) \\
r_t &= sigm(W_r \cdot x_t + U_r \cdot h_{t-1}) \\
\tilde{h}_t &= tanh\big(W_h \cdot x_t + U_h \cdot (r_t * h_{t-1})\big) \\
h_t &= (1 - z_t) * h_{t-1} + z_t * \tilde{h}_t
\end{aligned} \quad (1)
$$

GRU architectures typically employ multiple stacked layers to capture complex hierarchical patterns in time series data. Each additional layer allows the model to learn increasingly abstract representations of the temporal dynamics. However, in our specific application, which focuses on modeling monthly historical time series, much of the long-term dynamics are already explained by exogenous. Given this characteristic, we simplify the architecture by applying only a single GRU layer. This streamlined approach reduces computational complexity and minimizes the risk of overfitting while still effectively capturing the relevant temporal patterns inherent in the data.

The input *x(t)* to the architecture is composed of the combination of precipitation and temperature grids over the basin area. These grids represent the spatial distribution of precipitation and temperature, which are crucial variables influencing river discharge in each basin. The hydrological processes that govern the discharge of rivers are inherently linked to the drainage areas of the basin. These areas capture the precipitation that falls at various points across the basin, leading to the flow of water into the rivers that feed the hydroelectric plants within the basin.

By incorporating the full spatial distribution of precipitation and temperature across the entire basin, the model allows for the automatic discovery of regions within the basin that have the most significant impact on the river discharges. This is in contrast to traditional methods that aggregate these variables, for instance by using the mean precipitation over the entire basin.

On the other hand, using the full grid representation as input introduces a large number of variables, which can lead to overfitting if not properly addressed. With many input features, the model may end up learning not only the relevant patterns but also noise or spurious relationships in the data, resulting in poor generalization to new, unseen data.

To mitigate this issue, the proposed architecture combines a classic deep learning technique for handling overfitting—namely, the use of dropout layers—with an additional approach that makes use of the physical knowledge of the relationship between the input variables (particularly precipitation) and the output (discharge).

Dropout layers are a regularization technique used to prevent overfitting by randomly deactivating a fraction of the neurons during training. This randomness forces the model to learn more robust and generalized features, as it cannot rely on any particular subset of the input features. By randomly "dropping out" neurons during each forward pass, the model is encouraged to develop redundant representations of the data, which helps improve its ability to generalize to unseen data.

In Fig.4, an example of the dropout layers process is presented.

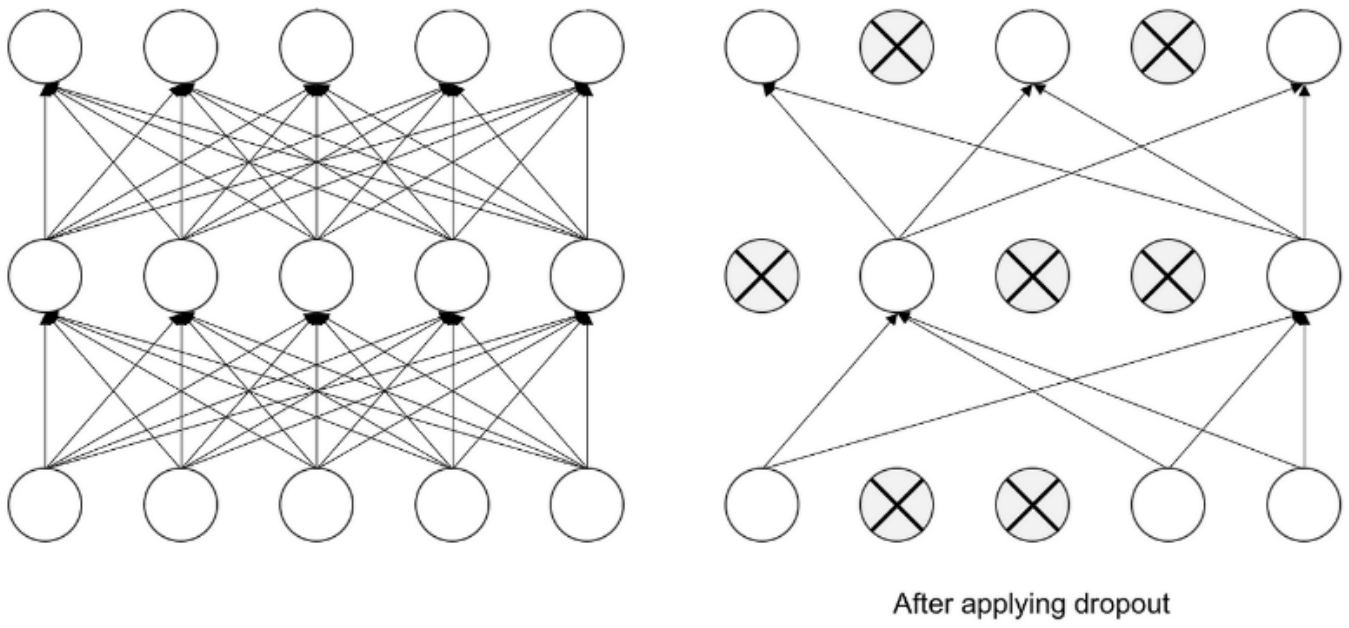

Fig.4. Dropout Layers [9]

In the context of our model, dropout layers help ensure that the relationships learned from the input grid of precipitation and temperature do not overly depend on specific locations or features, improving the robustness of the discharge predictions across different regions.

To complement the regularization provided by dropout layers, the proposed architecture incorporates an additional mechanism that enforces a critical physical constraint: the parameters learned by the network must ensure that the relationship between the output discharges and the precipitation inputs at time $t$ is always non-negative. In other words, a positive variation in precipitation at any grid point must never result in a negative variation, of any magnitude, in the corresponding river discharges.

If this type of constraint is not enforced, the large number of input variables introduces a high risk of spurious correlations being learned from the historical data. These correlations can result in the network assigning negative weights to certain grid points, where variations in precipitation are erroneously modeled as having a negative influence on river discharges. While this might superficially reduce the training error, it often comes at the cost of severely compromising the generalization capability of the model.

Operationalizing this constraint in practice requires addressing the fact that most deep learning frameworks do not natively support explicit constraints on the learned parameters. To overcome this limitation, alternative strategies must be implemented. In this work, we use the functionality of the *PyTorch framework* [10] to create a *callback* mechanism that modifies the model weights after each optimization step. This is achieved by implementing a custom *callback* that systematically enforces the non-negativity constraint on specific layers of the architecture.

In Fig.5 illustrates the implemented scheme, highlighting the custom module *NonNegativeLinear*, which extends the fully connected linear layer by incorporating a method to clamp the weights, ensuring they remain non-negative. The figure also shows the steps during which the callback is invoked within the process.

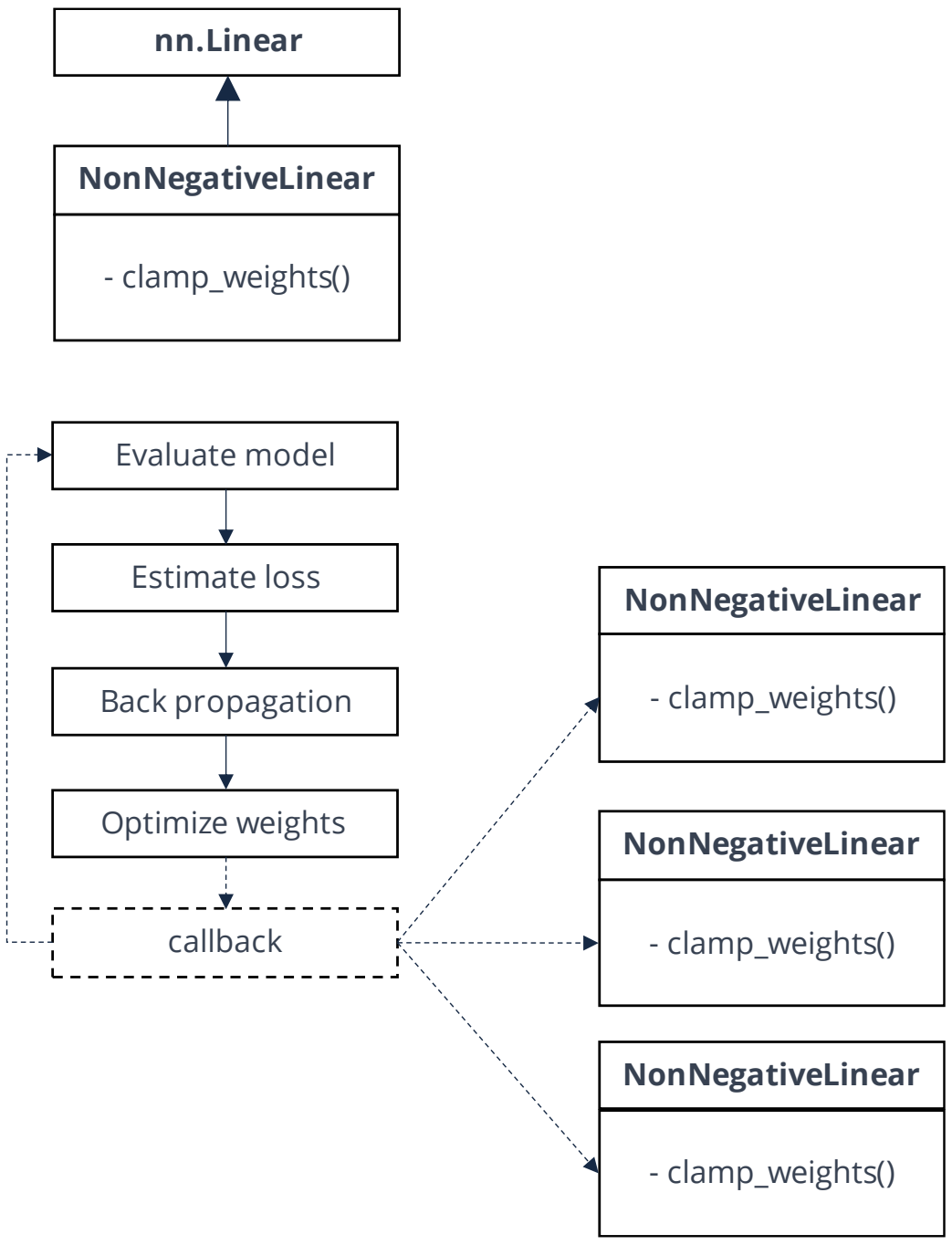

Fig.5. Custom Non Negative Layer and Callback

After each optimization step, the *callback* inspects the weights of this layer and adjusts any negative values to zero, thereby ensuring that all learned parameters respect the physical constraint of non-negativity. This approach not only integrates seamlessly into the training process but also guarantees that the resulting model maintains a physically meaningful representation of the input-output relationship at every stage of optimization.

### *B. Probabilistic Output Layer*

The proposed architecture introduces a probabilistic output layer as its final component, setting it apart from traditional recurrent architectures. Instead of generating deterministic predictions for the target time series at time *t*, this layer outputs the parameters of a three-parameter log-normal distribution: $\mu$, $\sigma$, and $\theta$. These parameters define the conditional probability distribution of the target variable at each time step, capturing not only the expected value but also the uncertainty and asymmetry inherent in the data. The functional form of this distribution is given by (2):

$$f(y) = exp(\mu + \sigma y) + \theta \quad (2)$$

Where $\theta$ represents a location parameter, $\mu$ and $\sigma$ control the shape and scale of the distribution.

The probabilistic output layer enables the architecture to effectively capture the stochastic nature of the outputs, providing a more comprehensive representation of the variability. This approach is particularly well-suited for long term hydrological scenario generation, where quantifying the range of possible outcomes is as critical as predicting the expected values. The use of the three-parameter log-normal distribution further strengthens this approach, as it is a widely adopted method for representing the uncertainty of positive-valued variables like river discharges. Its suitability arises from its ability to effectively model the characteristic skewness and heavy-tailed behavior often observed in such data, making it a natural fit for this domain.

Another significant advantage of the probabilistic output layer is its ability to estimate conditional probability distributions directly, based on the temporal and exogenous inputs. By conditioning the distribution parameters on the hidden states of the recurrent layer and external drivers like weather forecasts, the model directly integrates the influence of past dynamics and these critical exogenous inputs.

At each time step *t*, the output of the GRU layer passes through additional linear layers to obtain the terms that characterize the probability distribution function, as described in (3).

$$\begin{aligned} \mu_t &= W_\mu \cdot h_t \\ \sigma_t &= W_\sigma \cdot h_t \\ \theta_t &= W_\theta \cdot h_t \end{aligned} \quad (3)$$

Another significant shift in the proposed architecture lies in the choice of evaluation criterion used to construct the loss functions for training. Unlike traditional approaches, which typically rely on minimizing mean error or mean squared error, this architecture uses quantile regression as the foundation for its loss function. The primary motivation for this shift is to ensure that the model is not solely optimized for point estimates, such as the mean or mode, but is instead capable of accurately capturing the entire spectrum of the river discharge distribution. By focusing on specific quantiles, the architecture can directly optimize for the variability in the data, ensuring that the model performs well across the entire range of the discharge distribution. This approach avoids the pitfalls of traditional loss functions, which can emphasize the central tendency of the data and fail to account for the skewed and heavy-tailed nature of river discharge distributions.

The choice of a three-parameter log-normal distribution for characterizing the river discharge distribution offers a significant advantage, as the quantiles of this distribution can be efficiently expressed analytically as:

$$y_q = exp\left(\mu + \sigma \cdot \Phi^{-1}(q)\right) + \theta \quad (4)$$

Where $\Phi^{-1}(q)$ can be precomputed for each quantile *q:*

$$\Phi^{-1}(q) = \sqrt{2} \cdot erfinv(2q - 1) \quad (5)$$

For a set of quantiles $\Omega_q$ to be monitored, the loss function can be written as:

$$\begin{gathered} err_q = y - y_q \\ loss = \sum_{q \in \Omega_q} max\left(q \cdot err_q, [q-1] \cdot err_q\right) \end{gathered} \quad (6)$$

While mean-based regression minimizes the sum of squared residuals to capture the central tendency, quantile regression minimizes a weighted sum of residuals, where the weights are defined by the quantile of interest $q$. Specifically, for a given quantile, the loss function assigns higher penalties to overestimations or underestimations depending on their direction relative to $q$. For instance, when predicting the 95th percentile, overestimations are penalized less heavily than underestimations, reflecting the asymmetry in the importance of errors for that quantile. The expectation is that the weighted loss function guides the architecture to find a model where approximately 95% of the observed values fall below the predicted 95th quantile, while the remaining 5% lie above it.

Fig.6 illustrates an example of quantile regression applied to a river dischard time series, focusing on the 0.10 (lower quantile) and 0.95 (upper quantile) levels. The adjusted upper quantile series demonstrates a consistent trend of remaining above the majority of observations, with the expectation that approximately 5% of the observed values will exceed it. Conversely, the lower quantile series aligns with the expectation that roughly 10% of the observed values will fall below it.

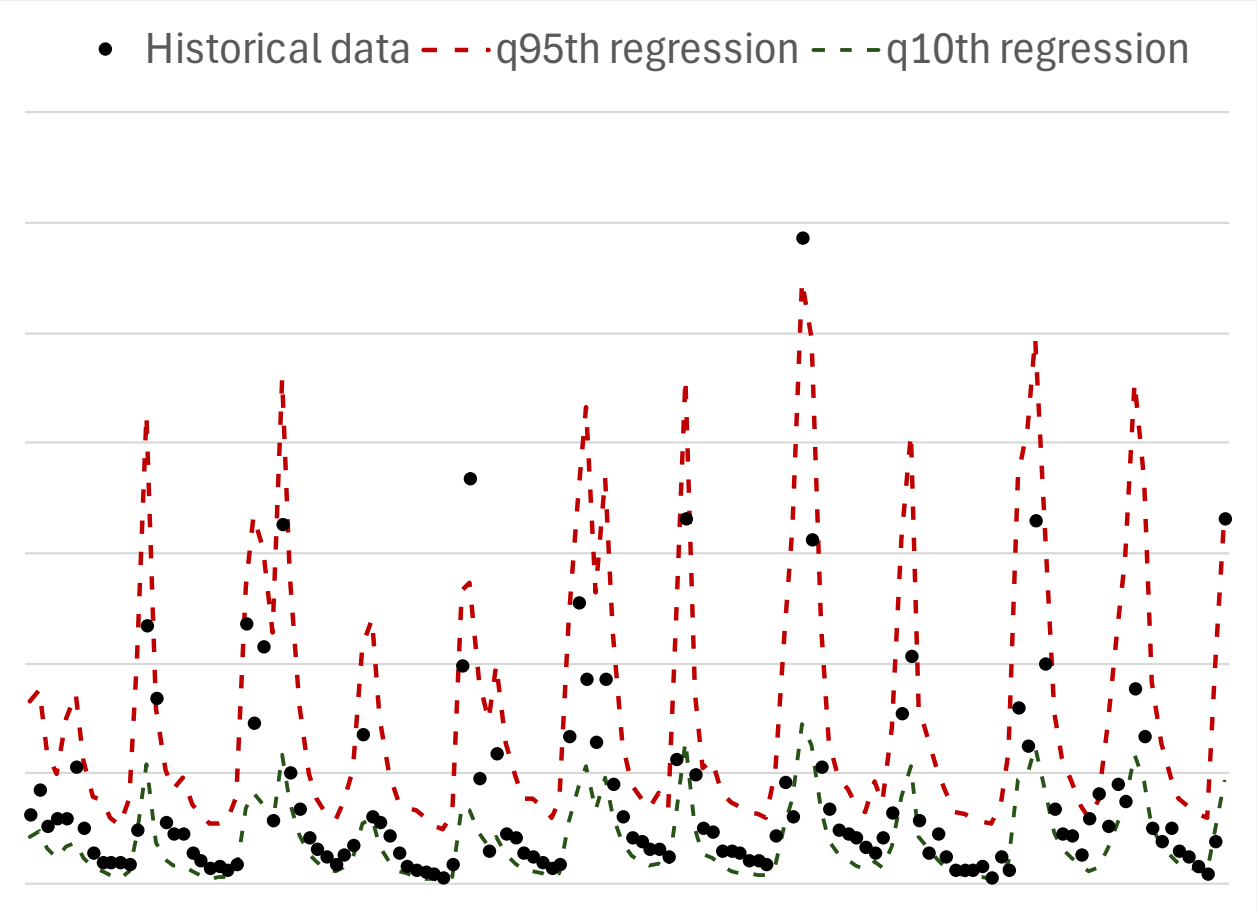

Fig.6. Quantile regression example

### C. *Serial Correlation*

The proposed method has an initial limitation compared to traditional statistical models for scenario generation: it does not inherently ensure the serial correlation of generated river discharge scenarios across consecutive stages. Statistical approaches, such as autoregressive models, generate scenarios directly, embedding the temporal dependence naturally into the outputs. In contrast, the proposed architecture produces the parameters of probabilistic distributions at each time step *t*, rather than the scenarios themselves. While this probabilistic framework effectively captures the uncertainty and variability at each stage, it lacks a direct mechanism to enforce temporal coherence in the generated scenarios.

It is important to note, however, that because the scenarios are conditioned on the trajectories of weather forecast ensembles, the temporal relationships provided by the climate model trajectories are indirectly incorporated. This ensures some level of temporal consistency across stages based on the external drivers. However, as the proposed method generates multiple hydrological scenarios for each given climate trajectory, the temporal evolution within each hydrological scenario is not inherently well-represented. This means that while the climate-driven variability is preserved, the serial correlation within the ensemble of discharge scenarios corresponding to a single climate trajectory requires further modeling.

To address the limitation of capturing serial correlation in the generated scenarios, an additional regression-based approach is introduced. This approach involves estimating linear models that represents the discharge at time *t*, *y(t)*, as a function of *y(t−1)*, the discharge from the previous stage, and the outputs of the recurrent layer from prior stages, *h(t−1)*, expressed as:

$$y(t) = \emptyset_y \cdot y(t-1) + \emptyset_h \cdot h(t-1) + \varepsilon \qquad (7)$$

Where the terms $\emptyset_y$ and $\emptyset_h$ are the regression coefficients that respectively explain the discharge at the current stage based on the discharge at the immediately preceding stage and the output of the recurrent layer at the previous stage, and $\varepsilon$ is the residues that can be characterized by $\Theta = cov(\varepsilon)$.

While the regression itself is linear, the inclusion of *h(t−1)* allows the model to indirectly capture nonlinear dependencies between *y(t)* and the sequence of previous discharges *y(t−2)*, *y(t−3)*, *y(t−4)*,... This is because *h(t−1)* encapsulates complex temporal dynamics learned by the recurrent layer. The intuitive idea behind this parametrization is that, since *h(t−1)* is the same for all the generated scenarios that are sampled from the same trajectory of climatic variable projections, it depends solely on the specific trajectory and not on the previously generated discharge values. In this way, the regression model captures the linear relationship between consecutive discharge samples within the same climate trajectory, while introducing an additional complexity that enables the model to differentiate between the behavior of discharge sequences across different climate trajectories.

The purpose of this regression is not to replace the probabilistic output layer of the architecture, which remains responsible for generating scenarios by characterizing the probability distributions at each stage. Instead, the regression provides a mechanism for determining the optimal ordering of the scenarios generated at stage *t*, aligning them with the scenarios from stage *t-1*. Given the regression models parameters, we can evaluate the likelihood of each possible sequence of sampled discharges between *t-1* and *t*, identifying the sequence that maximizes the coherence with the learned temporal dependencies by finding the permutation that maximizes the likelihood:

$$\pi^* = argmax\ \mathcal{L}\left(\pi | Y_t, Y_{t-1}, H_{t-1}, \emptyset_y, \emptyset_h, \Theta\right) \qquad (8)$$

Where the terms $Y_t$, $Y_{t-1}$and $H_{t-1}$ are the sequences of scenarios generated in *t* and *t-1*, per weather forecast ensemble, and π is the permutation that associates the scenarios in *t-1* and *t*.

Assuming that the residuals follow a multivariate normal distribution, it is possible to establish an association between the likelihood measure and the Mahalanobis distance, and the objective can be parametrized as given in (9). The detailed explanation of this association is beyond the scope of this work, but it can be found in some references like [11]

$$\pi^* = argmin \sum_{i=1}^{n} \left(Y_{t,\pi(i)} - [Y_{t-1,i}\emptyset_y + H_{t-1,i}\emptyset_h]\right)^T \Theta^{-1} \left(Y_{t,\pi(i)} - [Y_{t-1,i}\emptyset_y + H_{t-1,i}\emptyset_h]\right) \qquad (9)$$

## III. SIMULATION AND RESULTS

This section presents the simulations and results obtained to evaluate the proposed model. The Brazilian system was chosen as the case study due to its predominance in the country's energy matrix, accounting for over 60% of total energy production, comprising more than 200 large-scale hydroelectric plants, distributed across nearly the entire national territory.

The Brazilian energy system is not only a critical test case due to its high dependence on hydroelectricity but also because the country is significantly influenced by various climatic factors. These include phenomena such as El Niño, the Atlantic Multidecadal Oscillation, and other long-term climatic cycles, all of which have a direct impact on the country's hydrology. These climatic events can lead to significant fluctuations in river discharge and water availability, which in turn affect hydroelectric generation

For energy planning and analysis, the Brazilian system is traditionally divided into four subsystems: *SUDESTE* (Southeast), *SUL* (South), *NORDESTE* (Northeast), and *NORTE* (North). Notably, the Central-West region of Brazil is represented within the *SUDESTE* subsystem.

In Fig.7, the annual aggregated percentile of inflow energy for each system is presented for the period from 2019 to 2023. The percentile refers to the average inflow energy over the entire historical period. That is, if in 2019 the *SUDESTE* percentile of inflow energy was approximately 80%, it means that the observed aggregated value corresponds to a value 20% below the historical average.

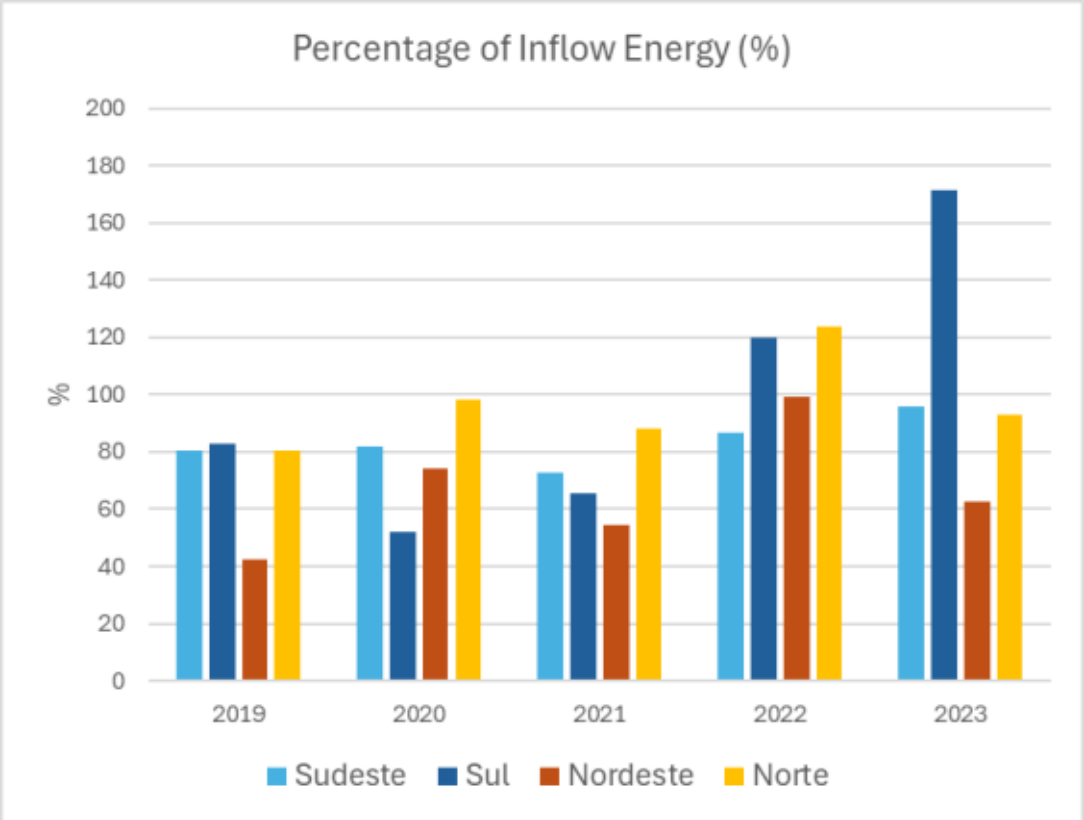

Fig.7. Percentage of Inflow Energy for the Subsystems of the Brazilian system

The analysis of the figure highlights the significant variability and unpredictability of hydrology. It is particularly noticeable that several consecutive years of unfavorable hydrology, especially in the subsystem *NORDESTE*, contribute to a trend of inflows remaining below historical averages. Additionally, the subsystem *SUL* displays high variability, further emphasizing the challenges in forecasting and managing hydropower production within the Brazilian system.

This results section is structured into two parts. The first part evaluates the model's ability to generalize effectively to climate inputs, focusing on its performance without yet incorporating scenarios generated by GCM models. The second part then examines the model's response when these GCM-generated scenarios are applied.

### *A. Model Validation*

The model will be trained using grid-based precipitation data from the CPC Gauge-Based Analysis of Global Daily Precipitation (CPC-Global)[12]. It is a dataset developed by NOAA's Climate Prediction Center (NOAA/CPC) under the CPC Unified Precipitation Project, that combines various sources of precipitation data using interpolation techniques to produce a unified and robust dataset. Daily precipitation data is collected from approximately 30,000 surface stations worldwide, including data from the World Meteorological Organization, cooperative observation networks, and national meteorological agencies.

The selection of the most suitable precipitation dataset is an important subject of study in its own right. However, for this work, the CPC-Global dataset was initially chosen due to its accessibility, daily updates, and the relatively minor differences in performance observed among datasets when evaluated in recent literature. A more detailed evaluation of the best model for Brazil, or other countries, could be the focus of future work.

In this initial analysis, the model is trained on data from 1981 to 2018 and evaluated on its ability to generalize the scenario generation process for the period 2019–2023. To ensure an unbiased assessment, the projections are made using an out-of-sample approach, meaning data from the validation period is excluded from the training phase.

The quantile regression loss function as adjusted to estimate $10^{th}$, $25^{th}$, $60^{th}$ and $95^{th}$ percentiles. This selection ensures a robust representation of both the lower and upper extremes, as well as the regions that are typically the most probable within a lognormal distribution.

In Fig.8, the results for four individual power plants are presented, each representing the subsystem to which it belongs, with the results discretized on a monthly basis. From top to bottom, we have the *ITAIPU* plant, representing the *SUDESTE* subsystem; the *ITA* plant, representing the *SUL* subsystem; the *SOBRADINHO* plant, representing the *NORDESTE* subsystem; and the *TUCURUI* plant, representing the *NORTE* subsystem.

Although the training was performed using data from 1981 onwards, only the series from 2010 are presented for greater clarity in the graphs. The black dashed line represents the observed historical data, while the areas with two distinct shades of red correspond to the limits of the lower and upper quantiles. The blue dashed line marks the boundary between the training horizon and the model evaluation horizon.

In Fig.9, the same result is presented, but only for the validation region, to provide a better view of how the observed real data aligns with the ranges of the estimated distribution.

Although this initial analysis is purely qualitative, it is possible to observe that both during training and validation, the probability distribution closely follows the pattern of the real data, capturing both the seasonality and the variations around the monthly patterns.

Table 1 compares the theoretical probabilities of each time series falling within specific value ranges with the observed frequencies, calculated as the ratio of actual occurrences to the total size of the series. The table is divided into three distinct value ranges: the probability of the time series values falling within the most probable region, defined as the range between Q2 and Q3; the probability of values falling below the lower quantile Q1, representing the less probable lower extremes; and the probability of values exceeding the upper quantile Q4, representing the less probable upper extremes. For each range, the table presents the reference value, the estimated values based on the training period, and the values for the validation period.

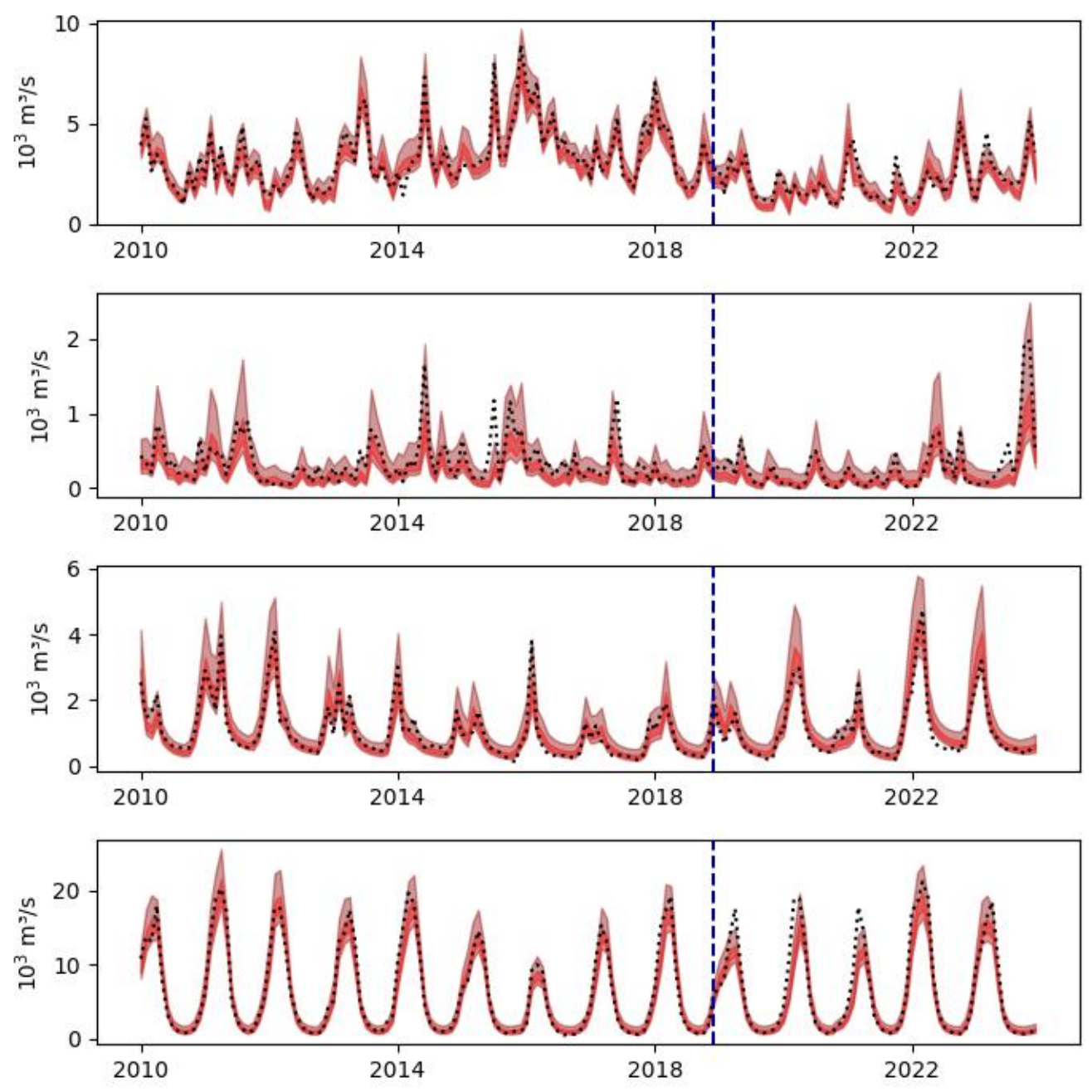

Fig.8. Monthly Results for Individual Plants

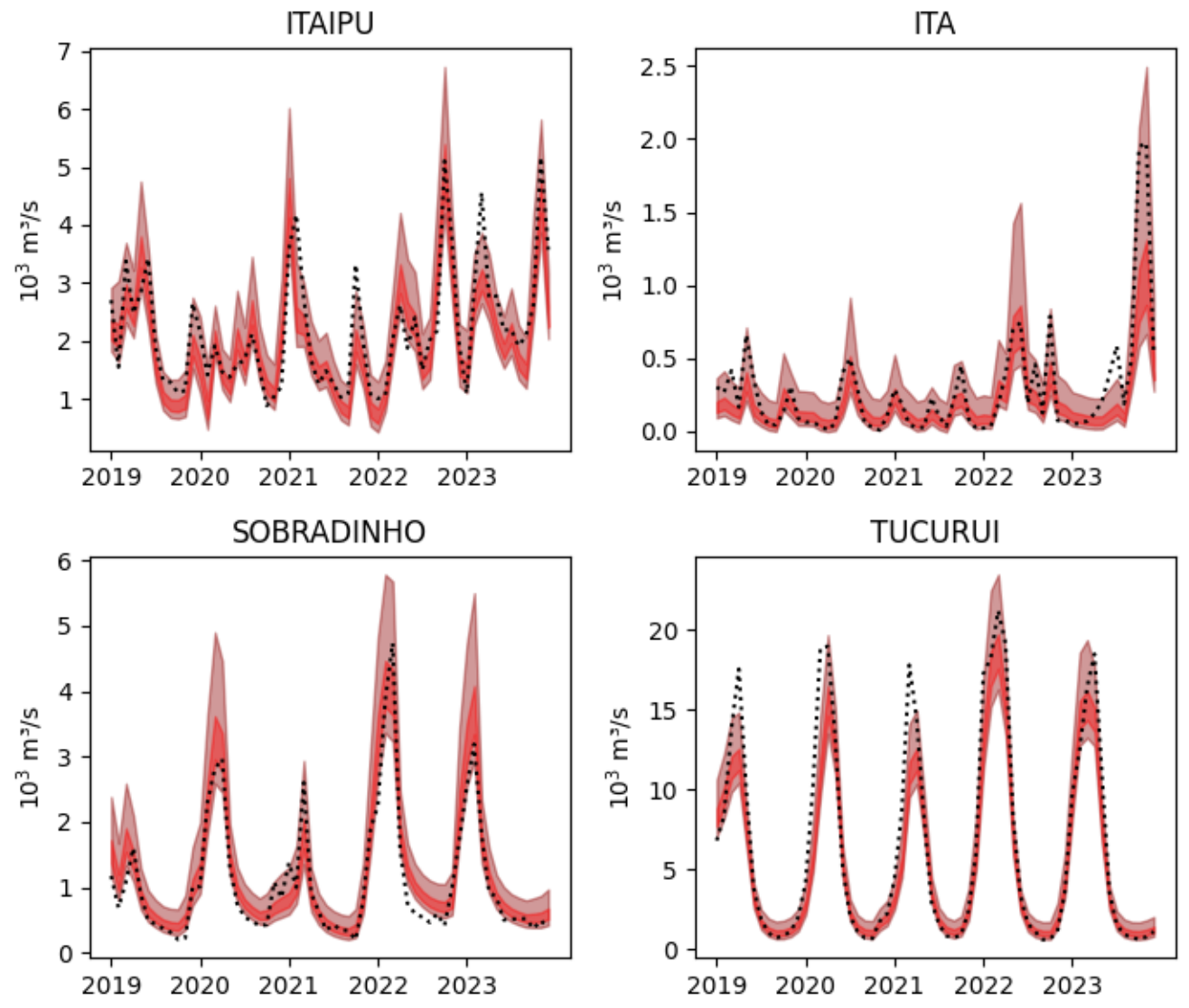

Fig.9. Monthly Results for Individual Plants (only validation period)

| Plant | Prob% ( Q2<X<Q3) | | | Prob% ( X<Q1) | | | Prob% ( X>Q4) | | |
|---|---|---|---|---|---|---|---|---|---|
| | Ref | Train | Valid | Ref | Train | Valid | Ref | Train | Valid |
| ITAIPU | 35 | 40 | 23.3 | 10 | 7.8 | 15 | 5 | 3.9 | 8.3 |
| ITA | 35 | 32.6 | 25 | 10 | 5.2 | 8.3 | 5 | 7.6 | 6.7 |
| SOBRADINHO | 35 | 39.2 | 33.3 | 10 | 11.5 | 33.3 | 5 | 1.4 | 1.7 |
| TUCURUI | 35 | 38.4 | 26.7 | 10 | 7 | 10 | 5 | 4.3 | 20 |

Table 1. Reference Probabilities vs Observed Frequencies

The number of occurrences for the most probable range showed strong consistency between the training and validation periods. However, for the extreme value ranges, discrepancies were observed in the validation data for the *SOBRADINHO* plant (lower quantile) and the *TUCURUI* plant (upper quantile), as highlighted in yellow in the table. However, it can be argued that, based on the visual inspection of the validation period for these two timeseries, the distances between the series and the extreme quantile ranges are not significantly large. Additionally, even under these conditions, the model's performance still shows a clear advantage over the traditional statistical approach, for which the distribution would tend to converge to the historical distribution. For instance, for *SOBRADINHO*, in the worst-case scenario (where the actual value is 203 m³/s), the 10% quantile would correspond to 337 m³/s, a value much higher than the one predicted by the model (266 m³/s).

For a more comprehensive analysis, the process of comparing probabilities for the same quantile ranges was applied to all hydroelectric power plants in the Brazilian system. In Fig.10, a 2D density plot is presented, showing the probabilities obtained on the *Y-axis*, compared to the reference probability (blue dashed line), considering only the validation period, but jointly considering the average discharge of each power plant, given on the *X-axis*. The average discharge is presented on a logarithmic scale, as the range of variation is very large.

For the 3 probability ranges evaluated, the densities show that the vast majority of the frequencies found are around the expected theoretical probability, as can be seen by the much darker shades of blue. At the same time, it is observed that for the largest flow series, there is never a significant difference. This is important, as the largest order of magnitude of the flows is directly related to the greater importance of the power plant to the system.

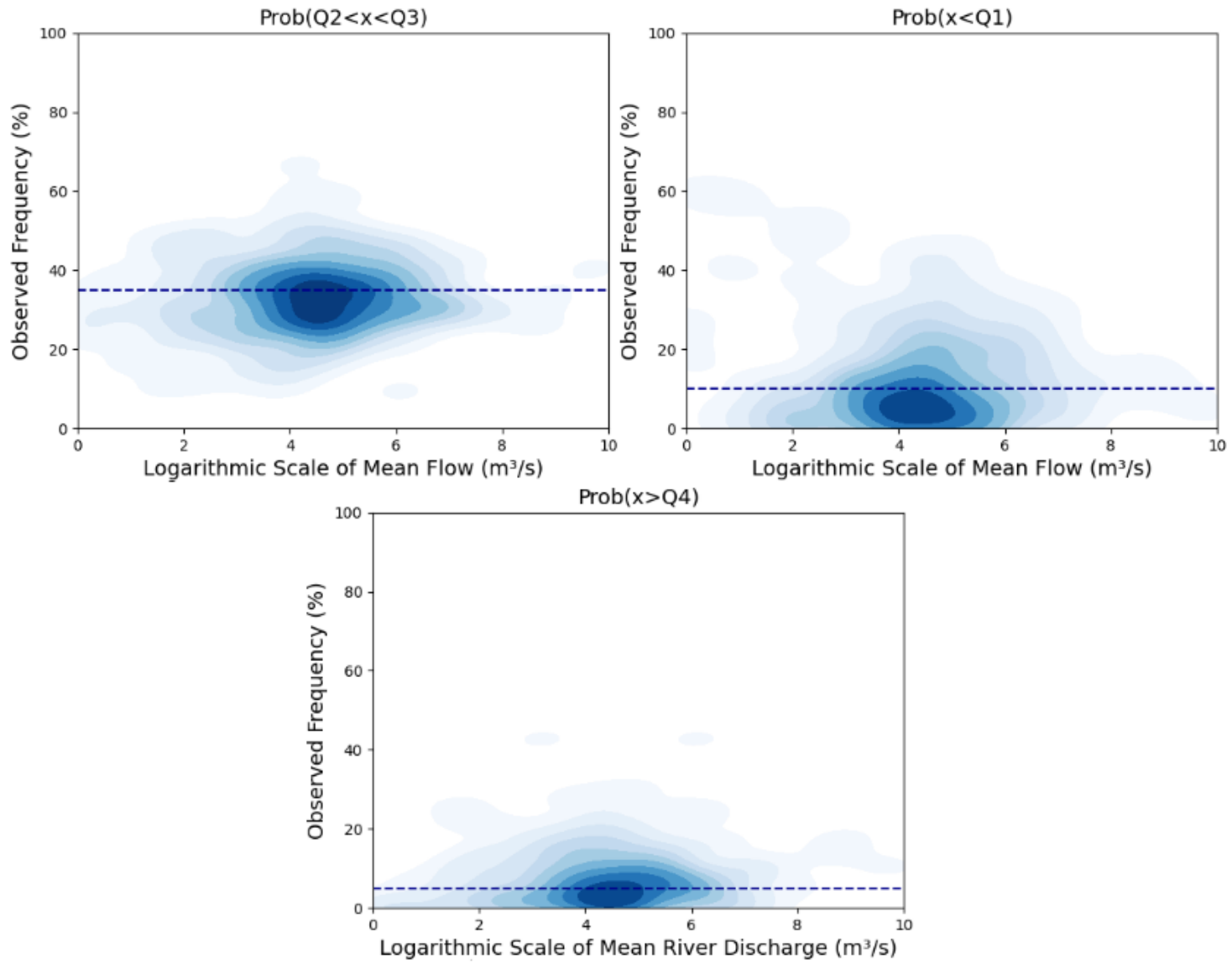

Fig.10. Density Estimate of Mean Discharge vs. Observed Frequency

For a more macro-level analysis, Fig.11 presents the aggregated results, focusing on the average annual inflow energy derived from historical river discharge and the corresponding scenarios generated by the proposed model. This type of macro-level analysis is particularly insightful because achieving satisfactory results in such aggregations requires the model to effectively capture both spatial and temporal correlations.

It is observed that for the four subsystems, most of the time, either the historical annual average energy lies within the most probable region of the distribution simulated by the model or is very close to it. Even when it falls outside this range, it remains within the second range, both in the training and validation intervals.

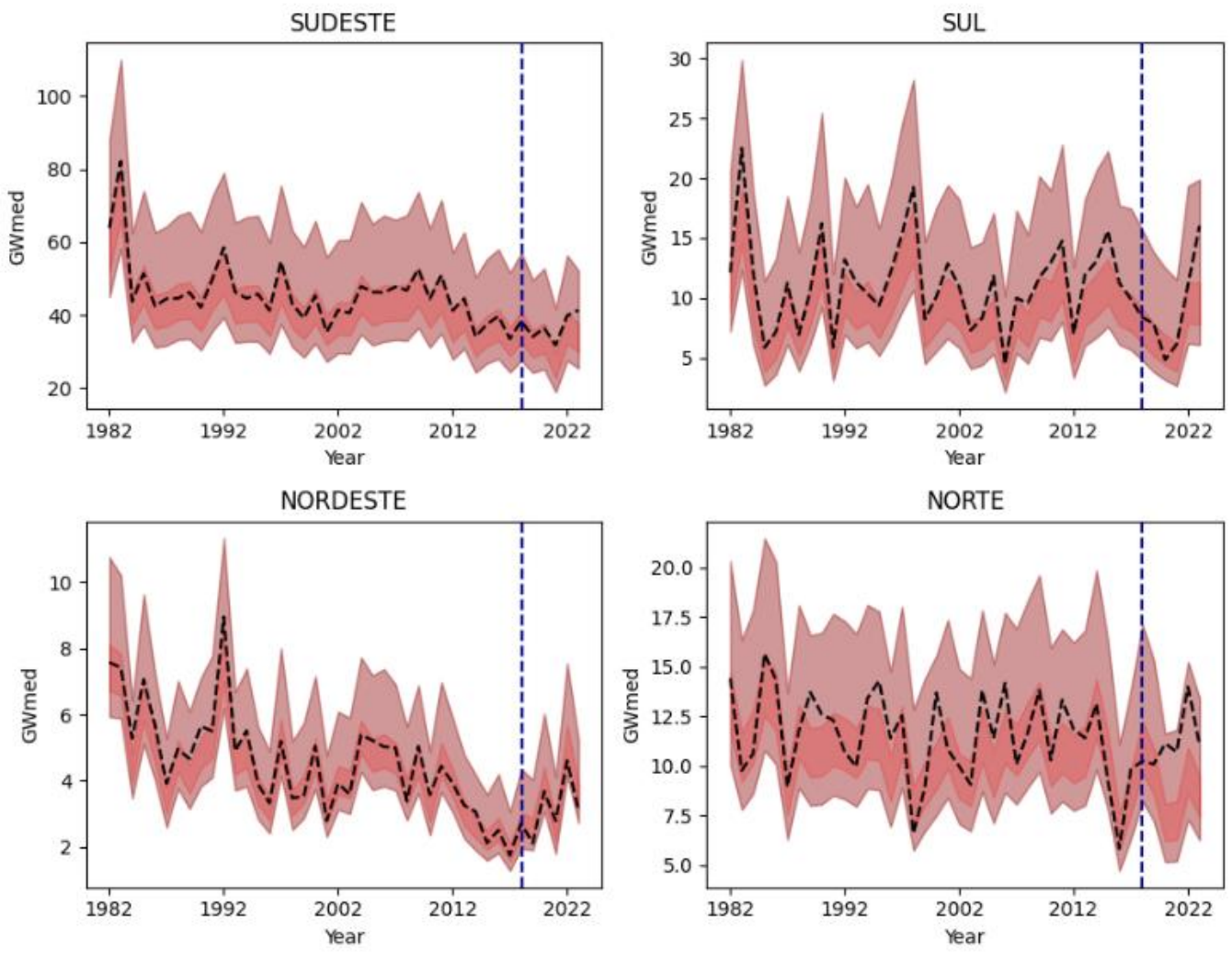

Fig.11. Annual Inflow Energy per System

## B. *GCM Ensemble Forecasts*

The adjusted model was then used to simulate the response of the Brazilian hydrology system based on ensemble forecasts from *SEAS5*[13], a GCM model from ECMWF. SEAS5 was chosen due to its extensive reference in the literature, particularly in the analysis of regions within Brazil and South America, for instance, [14]. It is important to note that the objective of this work was not to analyze the applicability of multiple and different GCM models. Instead, the focus was on evaluating the model's performance using a specific ensemble forecast. A future study will explore the comparative effectiveness of various GCMs and their impact on the model's performance, both when used individually and in combination.

The tested procedure involved simulating transitions between dry and wet periods, as well as wet to dry periods. For the dry-to-wet transition, simulations started in September for the intervals: 2018-2019, 2019-2020, 2020-2021, 2021-2022, and 2022-2023, resulting in 10 separate applications of the GCM model ensembles. For the wet-to-dry transition, simulations began in March for equivalent intervals. The projection obtained for each analyzed period consists of 51 trajectories of 6 months (ensembles). For each of these trajectories, the proposed model generates a set of 30 conditioned scenarios and combines them to calculate the statistics over the entire sample (monthly). These ensembles correspond exactly to past projections, meaning forecasts extending up to 6 months ahead from the starting point of each respective initial condition. This procedure effectively replicates scenario generation as if the model were operational during the defined periods.

Since the analyses for the various periods examined exhibit some similar behaviors, only a few key distributions obtained will be presented in the following analyses. The graphs correspond to the quantiles of the inflow energy for the four subsystems, compared to the historical data observed for the period. Additionally, the two blue lines indicate the reference for the 0.10 and 0.95 quantiles of the historical data, representing the extreme quantiles that would result from applying a traditional statistical model.

Fig.12 illustrates the transition from the dry period of 2019 to the wet period of 2020. The behavior observed in this graph shows significant similarity to what is seen in subsequent years. *SUDESTE* and *NORTE* subsystems are well captured by the model, with the actual historical series falling within or very close to the region of highest probability. The *NORDESTE* subsystem is also consistently near the region of highest probability, although it starts very dry, at the lower quantile limits. It is observed that, for most of the time, the observed data falls below the lower reference quantile, indicating a challenge for traditional statistical models to generate scenarios as dry as those observed during this period. The *SUL* subsystem also shows an adequate representation until the end of the projection period, where the system becomes much drier than predicted, resulting in proximity to the lower quantile.

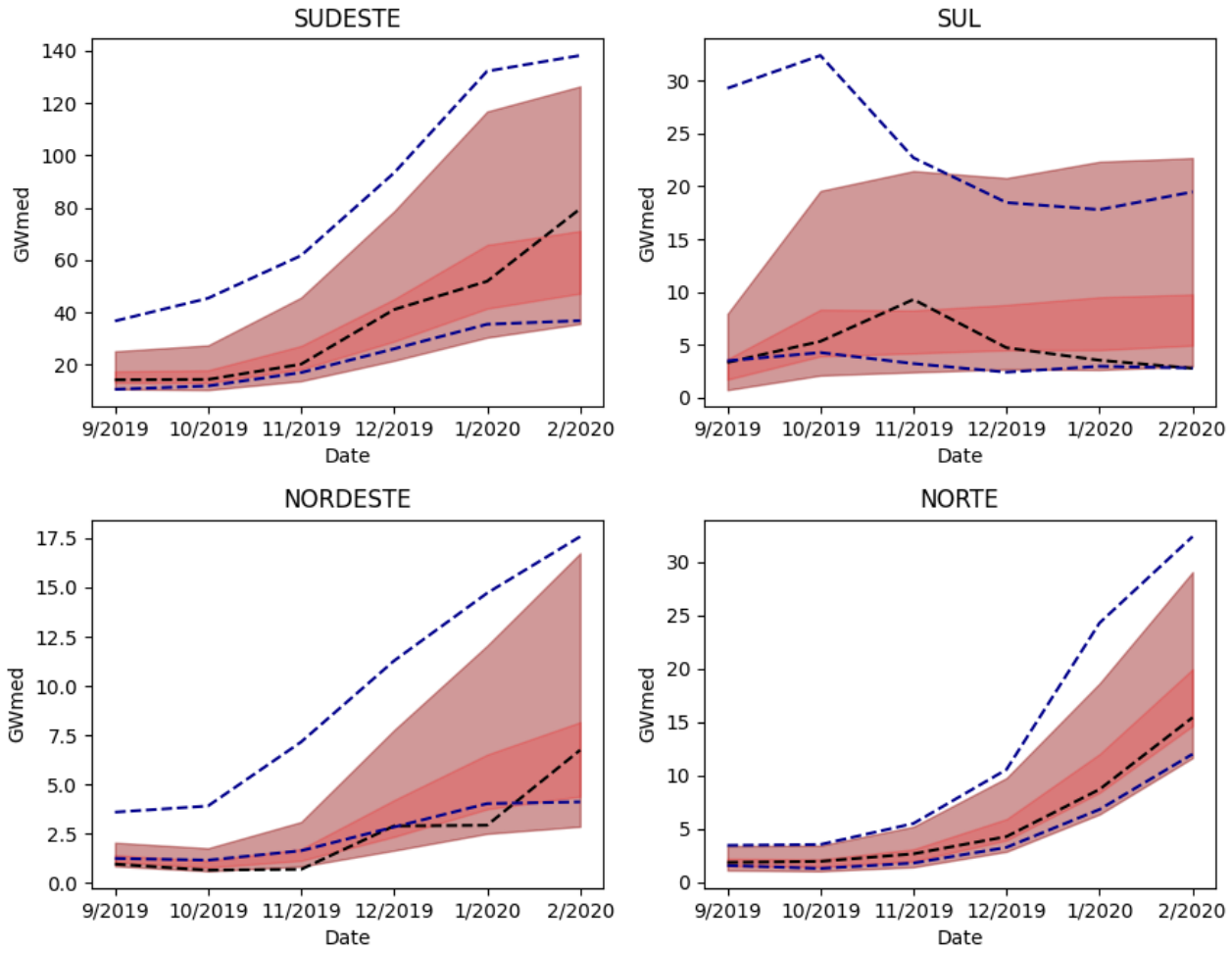

Fig.12. Projection Sep/2019-Feb/2020

Fig.13 presents the transition from the wet period to the dry period of the same year. Like the previous analysis, the characteristics observed for this year are also largely repeated in subsequent years. It is noticeable that the differences between the distributions captured by the model and the reference quantiles become significantly accentuated during this period, especially for the *NORDESTE* subsystem, but also for the early months in the *SUDESTE* subsystem. This indicates that traditional statistical models could generate scenarios much wetter than what was observed in practice. On the other hand, the proposed model generates highly representative scenarios for this period.

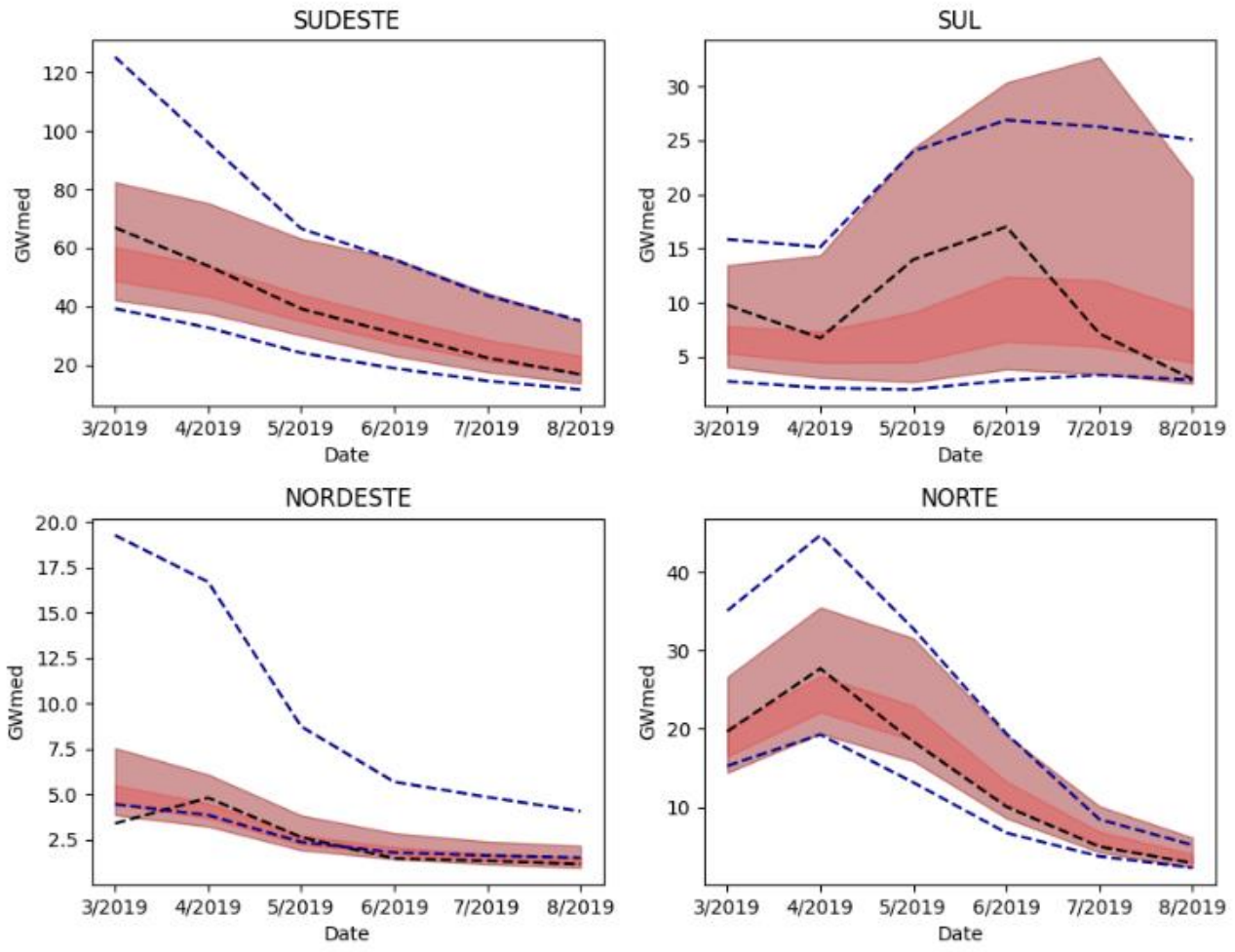

Fig.13. Projection Mar/2019 – Aug/2019

Fig.14 presents a configuration where, during the transition from the wet season of 2021 to the dry season, the actual inflow energy were much lower than those projected in advance, particularly for the *SUDESTE* subsystem and especially the *NORDESTE* subsystem. This resulted in a condition where the inflows fell below the model's lower quantile by June 2021. Nevertheless, the model successfully predicted that the distribution of inflows would be much drier than expected, as evidenced by comparing the results with the reference quantiles.

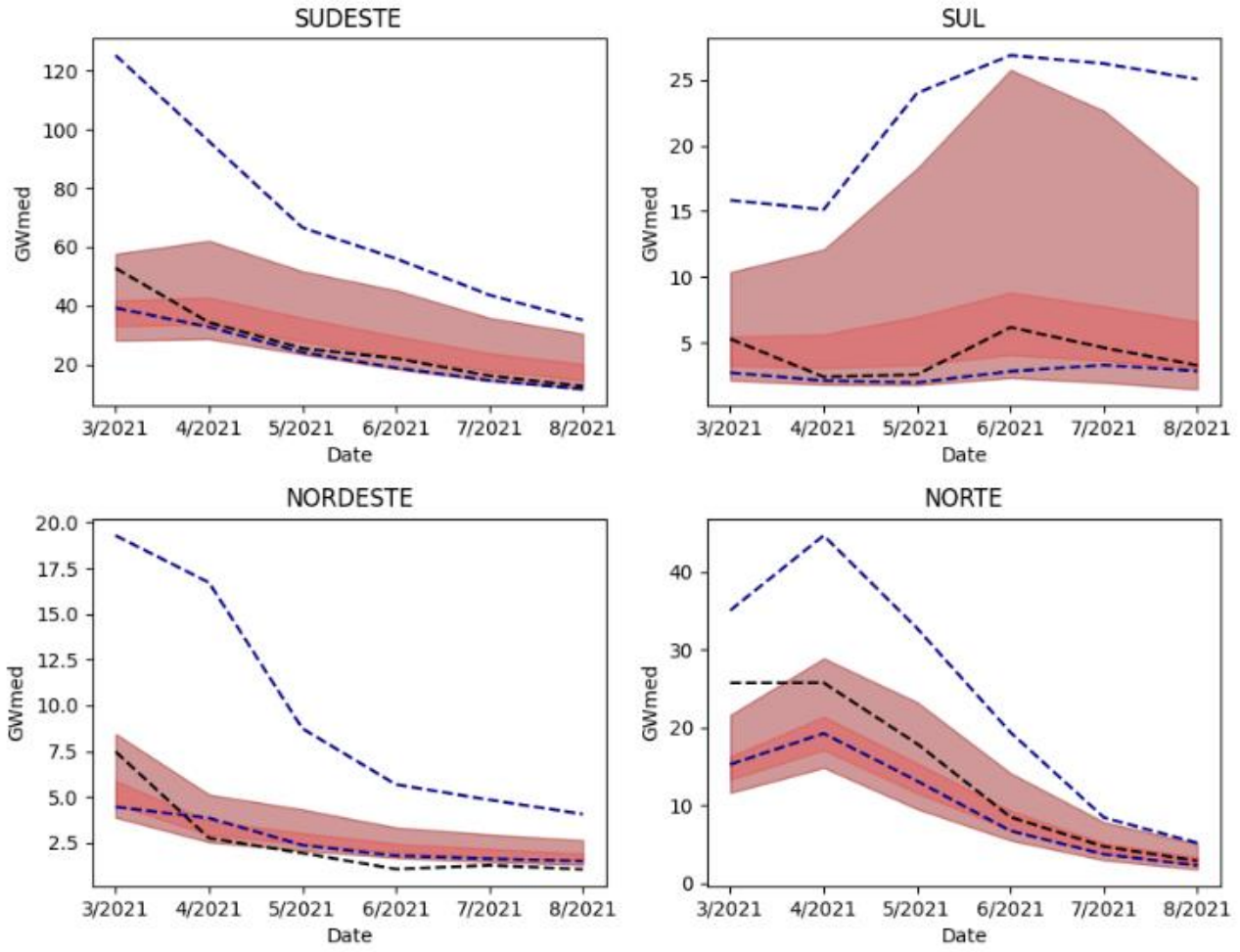

Fig.14. Projection Mar/2021-Aug/2021

However, in this case, it is interesting to note that in the projection for the following month, there was an adjustment in the SEAS5 expectations, so that the hydrological simulation now better reflects the observed real flow values.

Fig.15, which corresponds to simulating the model with the projection of the following month, i.e., for the period from April 2021 to September 2021, shows this characteristic, especially for the *NORDESTE* subsystem.

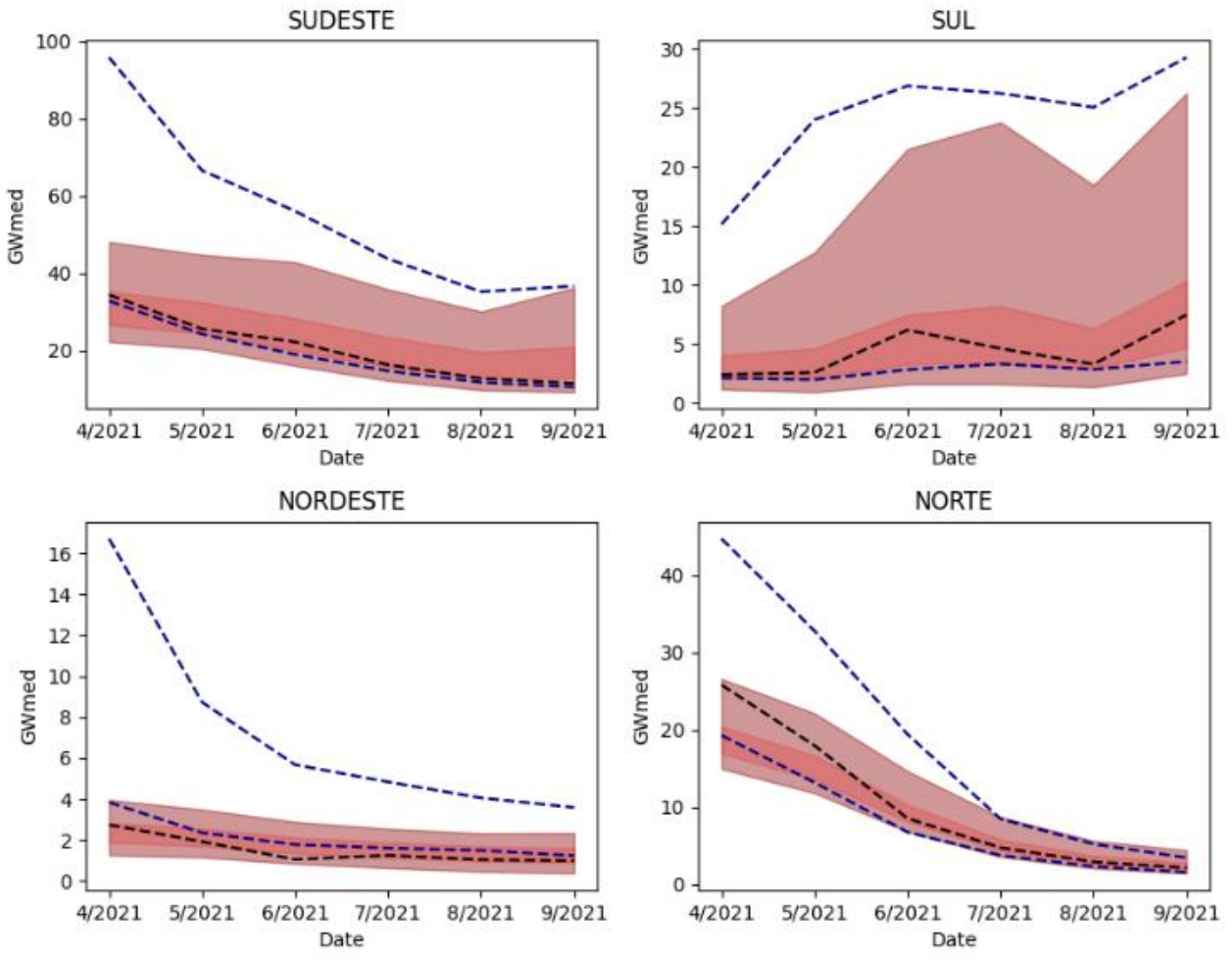

Fig.15. Projection Apr/2021-Sep/2021

## IV. CONCLUSION

This work proposed a deep learning-based model for generating long-term river discharge scenarios aimed at optimizing the operation of hydroelectric plants. The model is capable of integrating projections from General Circulation Models (GCMs) to enhance the plausibility of these scenarios. This is especially important in the context of hydrological modeling, where structural shifts in river flow patterns, driven by climate change and cyclic phenomena such as El Niño, are becoming increasingly significant. These changes can be better captured by GCMs, enabling the model to account for the complexities and uncertainties associated with future climatic variations.

To ensure the model's robustness despite the scarcity of historical data compared to the number of input variables (represented in a grid format), various approaches were proposed to avoid overfitting and overtraining, which is a common challenge in models with limited training datasets.

Simulations were conducted using the Brazilian hydroelectric system, and the results served both to validate the proposed architecture and to assess the model's performance when fed with projections from the SEAS5 GCM, provided by the ECMWF. The outcomes demonstrated the model's ability to generate realistic long-term discharge scenarios, with promising results in terms of the model's generalization capability and its response to GCM-driven input. Future studies will explore the comparative effectiveness of different GCMs and their impact on the model's performance.

## V. REFERENCES

[1] I. Pscheidt, A. M. Grimm. “The influence of El Niño and La Niña episodes on the frequency of extreme precipitation events in southern Brazil." Proc. Vol. 8. 2006.

[2] A. M. Grimm, "The El Niño impact on the summer monsoon in Brazil: regional processes versus remote influences." *Journal of Climate* 16.2: 263-280. 2003.

[3] R. R. Rodrigues, R. J. Haarsma, E. J. Campos, T. Ambrizzi. "The impacts of inter–El Niño variability on the tropical Atlantic and northeast Brazil climate". Journal of Climate, 24(13), 3402-3422. 2011.

[4] V. D. P. R. da Silva. "On climate variability in Northeast of Brazil. Journal of Arid Environments", 58(4), 575-596. 2004.

[5] J. A. Marengo, R. R. Torres, L. M.Alves. "Drought in Northeast Brazil—past, present, and future. Theoretical and Applied Climatology", 129, 1189-1200. 2017.

[6] P. H. Franses, R. Paap. "Forecasting with periodic autoregressive time series models". A Companion to Economic Forecasting, eds. MP Clements and DF Hendry, Oxford, Blackwell Publishers, 432-452. 2002.

[7] A. V. Vecchia. "PERIODIC AUTOREGRESSIVE-MOVING AVERAGE (PARMA) MODELING WITH APPLICATIONS TO WATER RESOURCES 1". JAWRA Journal of the American Water Resources Association, 21(5), 721-730. 1985.

[8] C. Kyunghyun; V. M. Bart; B. DZmitry; B. Fethi; S, Holger; B. Yoshua. “Learning Phrase Representations using RNN Encoder-Decoder for Statistical Machine Translation”. arXiv:1406.1078, 2014

[9] R. Shanmygamani,. “Deep Learning for Computer Vision”. 1. ed. Packt Publishing, 2018. ISBN 9781788295628.

[10] A.Paszke, S. Gross, F. Massa, A. Lerer, J. Bradbury, G. Chanan, S. Chintala. "PyTorch: An Imperative Style, High-Performance Deep Learning Library". In Advances in Neural Information Processing Systems 32 (pp. 8024–8035) 2019.

[11] J. Kuriakose, V. Amruth, G. Sandesh A., J. Venkata Naveenbabu, M. Shahid, & A. Shetty. (2014). Analysis of maximum likelihood and Mahalanobis distance for identifying cheating anchor nodes. arXiv e-prints, arXiv:1412.2857. https://doi.org/10.48550/arXiv.1412.2857.

[12] M. Chen, W. Shi, P. Xie, V. B. S. Silva, V. E. Kousky, R. W. Higgins, J. E. Janowiak, "Assessing objective techniques for gauge-based analyses of global daily precipitation". Journal of Geophysical Research Atmospheres, Washington, v. 113, p. 1-13, 2008.

[13] S. J. Johnson, T. N. Stockdale, L. Ferranti, M. A. Balmaseda, F. Molteni, L. Magnusson, S. Tietsche, D. Decremer, A. Weisheimer, G. Balsamo, S. P. E. Keeley, K. Mogensen, H. Zuo, & B. M. Monge-Sanz. (2019). SEAS5: The new ECMWF seasonal forecast system. Geoscientific Model Development, 12(3), 1087–1117. https://doi.org/10.5194/gmd-12-1087-2019.

[14] G. W. S. Ferreira, M. S. Reboita, & A. Drumond. (2022). Evaluation of ECMWF-SEAS5 seasonal temperature and precipitation predictions over South America. Climate, 10(9), 128. https://doi.org/10.3390/cli10090128.